\documentclass[letterpaper, 10 pt, conference]{ieeeconf}  

\IEEEoverridecommandlockouts               
\usepackage{cite}
\usepackage{graphicx}
\usepackage{amssymb}
\usepackage{hyperref}
\usepackage{amsmath,amsfonts} 

\usepackage{booktabs}

\usepackage{url}
\usepackage{float}
\usepackage{textcomp}

\usepackage{gensymb}
\usepackage{multicol,tabularx,capt-of}
\usepackage{multirow}

\usepackage{xcolor}
\definecolor{green}{rgb}{0,0.5,0}

\title{\LARGE \bf Flat'n'Fold: A Diverse Multi-Modal Dataset for Garment Perception and Manipulation
}

\author{Lipeng Zhuang$^{1}$, Shiyu Fan$^{1}$, Yingdong Ru$^{1}$, Florent Audonnet$^{1}$, \\ Paul Henderson$^{1}$ and Gerardo Aragon-Camarasa$^{1}$
\thanks{$^{1}$ School of Computing Science, University of Glasgow, G12 8QQ, Scotland, United Kingdom {\tt\small Lipeng.Zhuang@glasgow.ac.uk; gerardo.aragoncamarasa@glasgow.ac.uk}}}

\begin{document}

\maketitle
\thispagestyle{empty}
\pagestyle{empty}

\begin{abstract}

We present Flat'n'Fold, a novel large-scale dataset for garment manipulation that addresses critical gaps in existing datasets. Comprising 1,212 human and 887 robot demonstrations of flattening and folding 44 unique garments across 8 categories, Flat'n'Fold surpasses prior datasets in size, scope, and diversity. Our dataset uniquely captures the entire manipulation process from crumpled to folded states, providing synchronized multi-view RGB-D images, point clouds, and action data, including hand or gripper positions and rotations. We quantify the dataset's diversity and complexity compared to existing benchmarks and show that our dataset features natural and diverse manipulations of real-world demonstrations of human and robot demonstrations in terms of visual and action information. To showcase Flat'n'Fold's utility, we establish new benchmarks for grasping point prediction and subtask decomposition. Our evaluation of state-of-the-art models on these tasks reveals significant room for improvement. This underscores Flat'n'Fold's potential to drive advances in robotic perception and manipulation of deformable objects. Our dataset can be downloaded at \url{https://cvas-ug.github.io/flat-n-fold}

\end{abstract}

\section{Introduction}

Manipulating garments remains a significant challenge in robotics.
Tasks such as flattening and folding require understanding the vast space of configurations that garments can adopt~\cite{sun2015accurate,DBLP:journals/ijrr/VerleysenBw20}, and planning complex sequences of actions accordingly \cite{duan2022continuous,speedfolding2022}. This makes it vital to have large, diverse datasets from which to learn effective perception models and manipulation strategies. However, current efforts to capture a representative dataset of garments being manipulated \cite{DBLP:journals/ijrr/VerleysenBw20,DBLP:conf/cvpr/BozicZTN20,DBLP:journals/corr/abs-2203-11647} are limited in their utility for developing advanced robotic capabilities, due to their relatively small scope and diversity, as well as a lack of ground-truth action annotations.

In this paper, we introduce a novel dataset, \textit{Flat'n'Fold} (Fig. \ref{fig:process}), which includes 1,212 human and 887 robot demonstration sequences across 44 unique garments within eight categories and surpasses existing datasets in size, scope, and diversity (Tab. \ref{tab:Compare}). Flat'n'Fold is a multi-view, multi-modal dataset that captures the process of flattening and folding different garments, starting from a crumpled garment. Specifically, our dataset includes:
\begin{itemize}
\item Human Demonstrations, where twenty human participants performed the flattening and folding tasks. Our objective is to capture a wide array of human strategies in garment handling and the diversity and complexity of garment manipulation tasks.
\item Human-controlled Robot Demonstrations, where an expert human operator controls a robot to execute similar garment manipulation tasks, aiming to replicate natural, human-like approaches within the robot's operational limitations.
\end{itemize}
Moreover, our dataset is the first to include synchronized RGB-D image sequences with the position and rotation of the participants' hands (in human demonstrations) and robot grippers (in robot demonstrations), as well as camera parameters. To show the utility of our dataset, we set new benchmarks and baseline metrics for:

\begin{figure}[t]
   \centering
   \includegraphics[width=\linewidth]{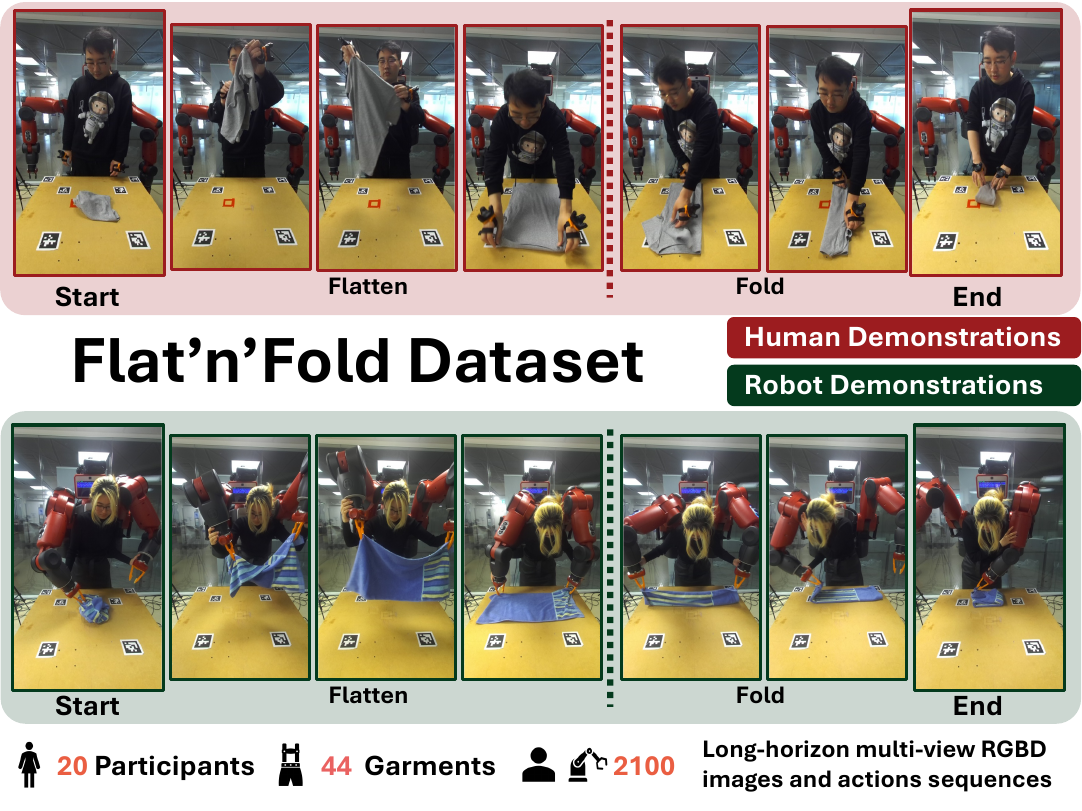}
   \vspace{-0.5cm}
   \caption{Human and robot demonstrations: Each demonstrates the progression from a crumpled garment configuration to a flattened and folded state.}\label{fig:process}
\end{figure}

\begin{itemize}

\item \textit{Grasping point detection} with 6,329 annotated point clouds from the human demonstrations and 5,574 from the robot demonstrations dataset. For this, we evaluated two popular models, PointNet++ and Point-BERT, and established a performance baseline.

\item \textit{Subtask decomposition from human and robot demonstrations} with $\sim$20,000 annotated sub-task boundaries corresponding to `pick' and `place' actions.


\end{itemize}




\section{Related work}

\begin{table*}
\centering
\caption{Comparative Analysis with Other Datasets, * indicates that the subset related to deformable objects (or garments) is considered.}
\resizebox{1\textwidth}{!}{
\setlength{\tabcolsep}{2pt}
\begin{tabular}{@{}lccccccccccccc@{}}
\toprule
Dataset &  Agent  & Setting      & Garment Cat.        & \#Garments    &\#Seq. & \#Frame       & Deform. & Task     & Vis. Info & Hum. Action  & Ann. & Total Hrs.           \\  \midrule

VideoFolding~\cite{DBLP:journals/ijrr/VerleysenBw20} & Human      & Real        & 3          & -        & 1000 & 304k & Large & Flattening \& Folding  & Multi-view RGBD & x   & Arm keypoint   &  8.5h \\

DeepDeform~\cite{DBLP:conf/cvpr/BozicZTN20} & Human  & Real        & 1         & $\sim$110 & 400  & 39k & Medium & Move  & Single-view RGBD & x  & Sparse keypoint   & - \\

Garment Tracking~\cite{DBLP:conf/cvpr/XueXZTLDYL23} & Human    & Virtual        & 4          & -   &-    & 790k & Large & Flatten\&Folding & Rendered Multi-view RGBD  & -& Garment \& Hand pose & -   \\

VR~\cite{DBLP:conf/ccia/Boix-GranellFT22} & Human     & Virtual        & 3     & 3         & 60         & - & Medium & Folding  & Virtual Reality & -  & -  & - \\ 
Semantic State Est.~\cite{DBLP:journals/corr/abs-2203-11647}\hspace{-4pt}& Human     & Real         & 1          & 18        & -       & 33.6K & Medium & Pick-up \& Folding  & RGB & x   & Cloth semantic state  & -\\ \midrule
Roboturk*~\cite{DBLP:conf/iros/MandlekarBSTGZG19} & Robot     & Real        & 3         & 3        & 920        & - & Medium & Flatten  & Multi-view RGBD & -   & -& $\sim$33h \\ \midrule

MIME*~\cite{DBLP:conf/corl/SharmaMPG18} & Human\&Robot     & Real         & 1          & $\sim$5         &$\sim$ 265      & - & Small & Wipe(no manipulation)  & Single-view RGBD & x   & -   &  -\\ 

\textbf{Flat'n'Fold (ours)} & \textbf{Human\&Robot}    & \textbf{Real}        & \textbf{8}         & \textbf{44}         & \textbf{2112}         & \textbf{$\sim$1600k} & \textbf{Large} & \textbf{Flattening \& Folding}  & \textbf{Multi-view RGBD \& Pointcloud} & \checkmark  & \textbf{Grasping point\&time}  & \textbf{$>$100h}\\ 
\bottomrule
\end{tabular}}
\label{tab:Compare}
\end{table*}


Human demonstrations significantly enhance robotic training by providing detailed examples of natural,
dexterous manipulations from which to learn (e.g. ~\cite{DBLP:conf/corl/WangFSZ0XZA23,robomimic2021,johns2021coarse-to-fine,speedfolding2022}), instead of defining inflexible manipulation strategies by hand \cite{sun2015accurate,duan2022continuous,duan2022data, DBLP:conf/humanoids/MolettaWWK23}. Human demonstrations have shown to be essential for long-horizon tasks for manipulating rigid and simple deformable objects \cite{DBLP:journals/corr/abs-2402-10329} and have enabled robots to anticipate future states and make decisions by optimizing policies throughout the task's duration \cite{DBLP:journals/corr/abs-2303-04137,ze20243d}.

Several existing datasets of human demonstrations of manipulating deformable objects have been used to train robotic systems for various tasks. However, these datasets are limited and have not let robots acquire all the skills necessary to manipulate garments, as shown in Fig. \ref{fig:process}. For example, the MIME dataset~\cite{DBLP:conf/corl/SharmaMPG18} provided video demonstrations of 20 tasks but focused on rigid objects. That is, only one task involved a deformable object, \textit{wiping with a cloth}, which was captured using single-view RGB-D data. DeepFashion~\cite{DBLP:conf/cvpr/LiuLQWT16} included only clothing and was intended for recognition tasks such as classification and attribute recognition rather than manipulation. Similarly, DeepDeform~\cite{DBLP:conf/cvpr/BozicZTN20} presented an approach for non-rigid reconstruction of dynamic objects using RGB-D cameras. Tzelepis \textit{et al.}~\cite{DBLP:journals/corr/abs-2203-11647} introduced a dataset of RGB images of human demonstrations while handling a tablecloth with the aim of semantic state estimation. However, the tasks considered were relatively simple (e.g.~lifting and diagonal fold) and lacked complex manipulation sequences. The Video Human Demonstration dataset~\cite{DBLP:journals/ijrr/VerleysenBw20} consisted of Kinect cameras mounted on a folding table, and 1,000 folding demonstrations were captured from multiple angles and participants. While this dataset provides RGB-D images and skeleton keypoints of human actions, it did not include 3D spatial position and rotation information (nor camera parameters that could be used to infer these), which are required for imitation learning and policy learning in robotics. 

Similarly, existing datasets relevant to deformable objects that use robot demonstrations or surrogate systems remain limited in scope and diversity. For instance, GarmentTracking~\cite{DBLP:conf/cvpr/XueXZTLDYL23} introduced the VR-Garment recording system, which enabled users to interact with virtual garment models via a VR interface. The VR-Folding dataset was introduced and included complex garment poses for tasks such as flattening and folding. However, the VR-Folding dataset was inherently bound to virtual reality (VR) environments, which introduced limitations in terms of its relevance to practical robotic operations in real-world scenarios. For real-world applications, Boix-Granell \textit{et al}.~\cite{DBLP:conf/ccia/Boix-GranellFT22} designed a Unity-based 3D platform using an HTC Vive Pro system for tablecloth manipulations only. Similarly, Moletta \textit{et al.}~\cite{DBLP:conf/humanoids/MolettaWWK23} attempted to automate cloth folding with a system that employed skeleton representations to generate folding plans. Although this helped replicate specific folding techniques, only three garment classes and fixed folding procedures were considered, reducing its adaptability to the diverse techniques required in real-world settings. Another significant work, the Roboturk dataset~\cite{DBLP:conf/iros/MandlekarBSTGZG19}, used a smartphone-based 6-DoF controller with cloud integration, and a robotic arm was used to smooth out garments such as hand towels, jeans, or t-shirts on a table. Roboturk captured these sequences from a single camera perspective, restricting its application for complex tasks that require dual-arm coordination or depth perception.

To overcome the limitations of existing datasets, we introduce Flat'n'Fold, a dataset comprising human and robot demonstrations, including various garment types, manipulation tasks, and garment flattening and folding sequences, providing visual and action information. 
Tab.~\ref{tab:Compare} compares our Flat'n'Fold dataset against existing datasets focused on human demonstrations and robotic manipulation of garments. In the table, `Agent' represents whether it is human demonstrations or robot demonstrations; `Hum. Action' means whether human action data is recorded and saved during human demonstrations; `Ann.' represents extra annotations. Flat'n'Fold has a clear advantage in terms of data volume, diversity, and modalities recorded.

\section{Materials and Methodology}

\subsection{Hardware}

\begin{figure}[t]
    \centering
    \includegraphics[width=0.95\linewidth]{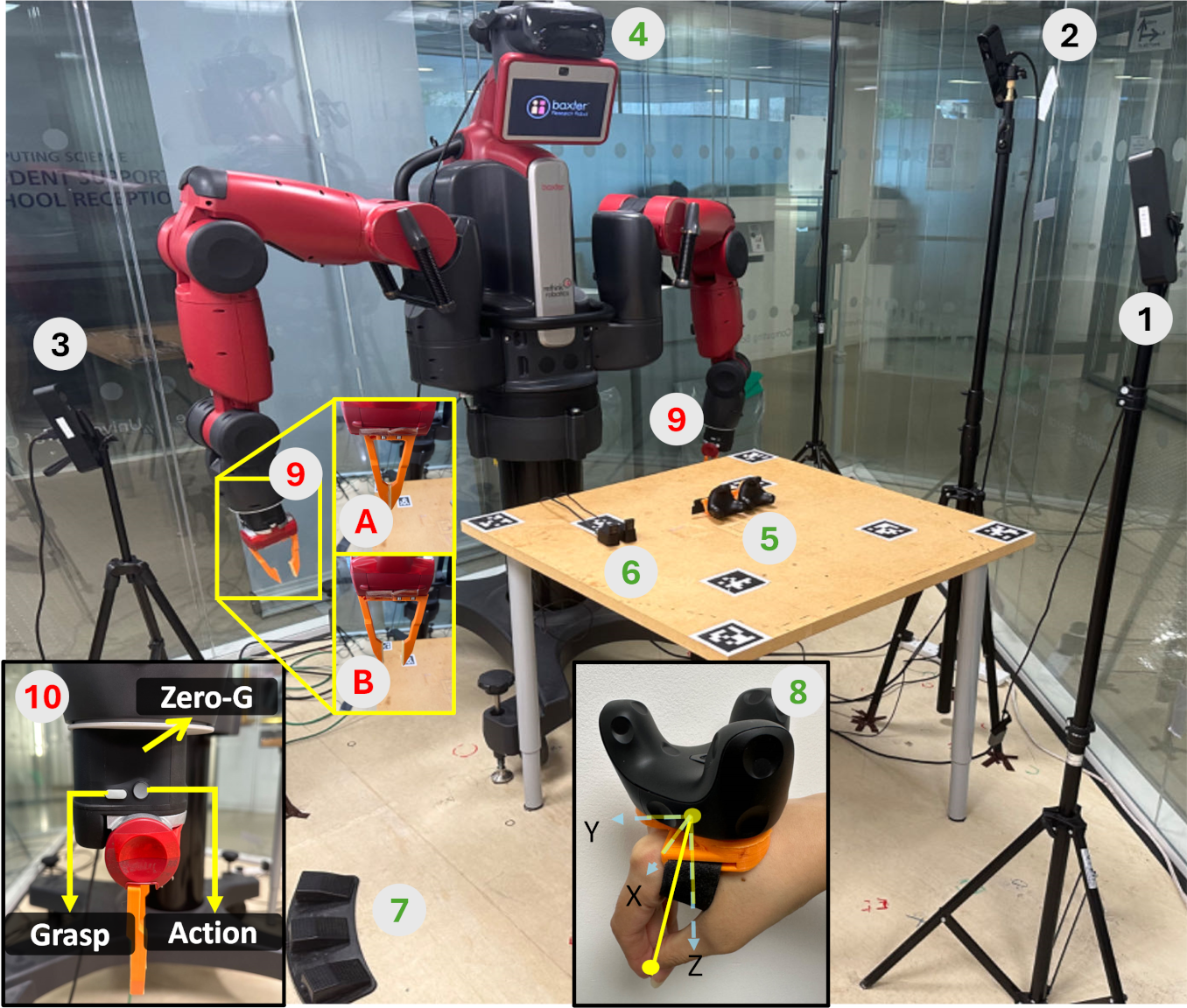}
    \caption{Hardware Setup. (1) Front camera; (2) Top camera; (3) Side camera; (4) Steam Index VR Headset~\cite{valveindex}, which serves as the origin of the world; (5) HTC Vive tracker; (6) Receiver of the tracker; (7) Pedal; (8) Grasping Point, the yellow line indicates the distance from the center of the tracker to the grasping point; (9) Baxter's gripper with (A) the gripper in its closed state and (B) opened state; (10) Baxter's zero-G mode~\cite{armcontrolsystem} and control buttons. The black numbers mean that this hardware was used for human and robot demonstrations; red numbers, it was used in the robot demonstration dataset only, and green numbers, it was used only for human demonstration. 
    }\label{fig:hardware}
\end{figure}

The hardware used is shown in Fig.~\ref{fig:hardware}. The setup for the human demonstration dataset integrates visual and motion information from three ZED2i cameras, a VR controller, and three mechanical pedals interfaced via USB. The ZED2i cameras were positioned to provide multiple views and minimize occlusions from human arms during garment manipulation. These views encompass the front, top, and back of the garment. The ZED2i cameras recorded 1080p images at 15Hz. For motion tracking, we used SteamVR~\cite{steamvr} with two HTC Vive Trackers, where the VR headset~\cite{valveindex} served as the world coordinate frame to which robot and tracker reference frames are linked. Two HTC Vive Trackers were affixed to the backs of participants’ hands, capturing position and rotation in 3D space relative to the headset at 15Hz. Three pedals were used to record the timing of actions involving the left or right hands and both hands, annotating when participants grasped or released the garment.

The setup for the robot dataset closely mirrored that for the human demonstrations. We used a Rethink Robotics Baxter robot equipped with custom 3D printed grippers\footnote{The STL files for the gripper used can be found at \url{https://cvas-ug.github.io/flat-n-fold}} for improved grip (Fig.~\ref{fig:hardware}). To ensure consistency across both datasets, we again used three ZED2i cameras~\ref{fig:hardware} with the same camera positions as in the human demonstrations. Control of the robot arm presented unique challenges due to the complexity and duration of garment manipulation tasks. To optimize the operator's speed, ease, maneuverability, control and the fidelity of the robot arm's movements to human actions, we explored direct teleoperation~\cite{audonnet2024telesim} and immersive Virtual Reality (VR) teleoperation \cite{audonnet2024immersim}. With the VR, the operator struggled with boundary detection of the garment due to resolution limitations. Teleoperation~\cite{audonnet2024telesim} was problematic as synchronous movement between the human hand and robotic arm obstructed the operator’s side view of the garment while repositioning the operator in front of the robot was impractical due to human joint movement limitations and the need to mirror the robot’s actions. To resolve these issues, we opted to directly manipulate Baxter's arms in its zero-G mode~\cite{armcontrolsystem}, (Fig.~\ref{fig:hardware}), which allows the operator to move the arms without resistance from the motors, as well as to activate the gripper. This mode was activated by grasping the cuff over its groove. The operator used a 'grasp' button alongside an 'action' button to control the opening and closing of the gripper; gripper opening and closing states are shown in Fig. \ref{fig:hardware}-A and B.

\subsection{ZED2i Camera Calibration}
We place calibration boards at various positions on the table (see Figs. \ref{fig:hardware} and \ref{fig:process}) for camera calibration and use OpenCV's PnP solver~\cite{DBLP:journals/cacm/FischlerB81} to estimate the relative camera pose with respect to the table. To establish the calibration board's position relative to the headset, a tracker was positioned in the right corner of each board (designated as the origin in OpenCV). 
The collected data was averaged over 10 repetitions to robustly estimate the relative positions of each camera to the headset, ensuring accurate spatial data in our experiments.

\subsection{Methodology}\label{sec:methodology}


For the human demonstration dataset, we invited 20 participants to capture how they manipulate garments from crumpled to folded states. Participants, equipped with trackers on both hands, were asked to stand behind a tablewere briefed on movement guideguidelines before data collection commenced. These guidelines included avoiding finger bending and only performing pick-and-place actions, minimizing arm crossing, and gently shaking the garment for easier handling. They were also advised to keep their hands within the bounds of the table as much as possible. For robot demonstrations, an operator familiar with the robotic arm controlled the robot's arms and grippers (Fig. \ref{fig:hardware}-9). 

The human and robot demonstration stages are illustrated in Fig.~\ref{fig:process}. For both, demonstrations began with the participants or robot operator grasping a crumpled garment from any point using either hand or gripper and lifting the garment from the table. The flattening stage was not standardized; hence, participants or robot operator grasped and manipulated the garment in the air until it was nearly flat and then laid on the table. After, they were allowed to make further adjustments to enhance the garment's flatness. We split participants in terms of whether they followed a fixed folding strategy (i.e. \textit{Fixed}) or their usual, natural way to fold garments in their daily life (i.e. \textit{Daily-life}). We must note that 7 participants performed both strategies; hence, the \textit{Fixed} strategy comprised 17 participants, while the \textit{Daily-life} strategy comprised 10 participants. The participants carrying out the \textit{Fixed} strategy adhered to a structured folding approach such that all participants had a similar folded garment at the end of the manipulation. This approach differs from the simpler folding methods used in previous studies and is intended to reflect realistic, practical folding techniques. The robot operator followed both strategies while folding garments.


\subsubsection{Data Format}

Both visual and action data were captured and stored using ROS2 bag files. We use nearest-timestep matching to synchronize images with action data (similar to ~\cite{DBLP:journals/corr/abs-2402-10329}). This synchronization ensures that for each camera, the number of RGB images equals the number of depth images. Moreover, the quantity of synchronized action data matches the number of images, ensuring that camera data aligns with the corresponding action data. 

In the human demonstration dataset, the action data comprises the position and rotation of the tracker's center. For all participants, we measured the distance from the back of their hand to the tip of their index finger along the three axes (Fig.~\ref{fig:hardware}-8). These measurements were then used to calculate the position of the grasping point, ensuring that we captured the location where participants interacted with the garment. A fixed offset representing the gripper length is added for the robot dataset, allowing for consistent alignment across both datasets.


\subsection{Dataset overview}



\begin{table}
\centering
\caption{Statistics of our dataset}
\label{tab:Dataset}
\resizebox{0.48\textwidth}{!}{%
\begin{tabular}{@{}lcccccc@{}}
\toprule
\multirow{2}{*}{\textbf{Garment type}} & \multicolumn{3}{c}{\textbf{Human demonstration}} & \multicolumn{3}{c}{\textbf{Human-controlled robot}} \\
\cmidrule(lr){2-4} \cmidrule(l){5-7} 
 & Fixed & Daily-life & Total & Fixed & Daily-life & Total \\
 \midrule
Napkin & 91 & 138 & 229 & 76 & 40 & 116 \\
Towel & 36 & 55 & 91 & 40 & 20 & 60 \\
LS-Shirt & 55 & 80 & 135 & 80 & 38 & 118\\
SS-Shirt & 66 & 99 & 165 & 80 & 37 & 117 \\
Pant & 56 & 80 & 136 & 79 & 40 & 119 \\
Sweater & 57 & 85 & 142 & 80 & 40 & 120 \\
LS-Tshirt & 61 & 91 & 152 & 78 & 40 & 118 \\
SS-Tshirt & 63 & 99 & 162 & 80 & 39 & 119 \\
\midrule
Total & 485 & 727 & 1212 & 593 & 294 & 887 \\
\bottomrule
\end{tabular}%
}
\end{table}

Our human and robot demonstration datasets use 44 diverse garments from 8 categories: napkins, towels, pants, long-sleeved shirts (LS-shirt), short-sleeved shirts (SS-shirt), long-sleeved t-shirts (LS-tshirt), short-sleeved t-shirts (SS-tshirt), and sweaters. Each category includes five items, except for pants, which consist of six. The garments within each category vary in size and stiffness, covering a wide range of fabrics and material properties. The dataset consists of 1,212 individual flattening and folding sequences of human and 887 robot demonstrations. The data distribution across different categories and individual garments is shown in Tab.~\ref{tab:Dataset}.

\subsubsection{Human Demonstration Dataset}

Each sequence comprises RGB images and depth images from three cameras, action sequences (capturing both the position and rotation of trackers related to the headset), and the timing of finger interactions with the garment. Each sequence documents a complete process of flattening and folding the garment. Additionally, each sequence records the position and rotation at the gripping points and is annotated with multiple labels: clothing type, clothing name, and the folding strategy of the action (\textit{Fixed} or \textit{Daily-life}). As described in Sec. \ref{sec:methodology}, the \textit{Fixed} strategy followed a standardized folding method, and the \textit{Daily-life} strategy followed a natural, everyday folding technique. 

\subsubsection{Robot Demonstration Dataset}
For the robot dataset, we collected 887 sequences using the same 44 garments as in the human demonstration dataset. 
This dataset comprises the robot's joint states and the position and rotation at the center of the wrist reference frame relative to its world origin, in addition to the multi-view RGB images and depth images. We also recorded gripper opening and closing states.

\section{Experiments}

We first compare the diversity and complexity of Flat'n'Fold to existing datasets (Sec.~\ref{sec:diversity_complexity}). 
Then, we define two benchmarks for evaluating grasp prediction (Sec.~\ref{sec:grasp_prediction}) and sub-task decomposition (Sec.~\ref{sec:subtask_decomposition}).

\begin{table*}[t]
\centering
\caption{Comparative analysis for action diversity.}
\label{tab:action_diversity}
\begin{tabular}{@{}lcccccccc@{}}
\toprule
\multirow{3}{*}{\textbf{Dataset}} & \multicolumn{4}{c}{\textbf{Human}} & \multicolumn{4}{c}{\textbf{Robot}} \\
\cmidrule{2-5}\cmidrule(l){6-9}
 & \multicolumn{2}{c}{\textbf{Average Complexity}} & \multicolumn{2}{c}{\textbf{Average Diversity}} & \multicolumn{2}{c}{\textbf{Average Complexity}} & \multicolumn{2}{c}{\textbf{Average Diversity}} \\
 \cmidrule{2-3}\cmidrule{4-5}\cmidrule(l){6-7}\cmidrule{8-9}
  & \textbf{Position}& \textbf{Rotation} & \textbf{Position}& \textbf{Rotation}&\textbf{Position} & \textbf{Rotation} & \textbf{Position}& \textbf{Rotation}\\
\midrule
Roboturk~\cite{DBLP:conf/iros/MandlekarBSTGZG19} & - & - & - & - & 0.105 & 14.907 & 0.123 & 16.564 \\
MIME~\cite{DBLP:conf/corl/SharmaMPG18} & - & - & - & - & $7.38 \times 10^{-3}$ & 12.335 & $7.21 \times 10^{-3}$ & 4.742 \\
Flat'n'Fold (ours) & 0.245 & 10.726 & 0.219 & 5.560 & 0.217 & 17.089 & 0.176 & 16.894 \\
\bottomrule
\end{tabular}

\vspace{2pt}
\textit{* Position is given in meters and rotation in radians}~~
\end{table*}

\subsection{Quantifying the diversity of the dataset}
\label{sec:diversity_complexity}

We first measure the diversity and complexity of our dataset compared to RoboTurk~\cite{DBLP:conf/iros/MandlekarBSTGZG19}, MIME~\cite{DBLP:conf/corl/SharmaMPG18}, VideoFolding~\cite{DBLP:journals/ijrr/VerleysenBw20}, and Semantic State Estimation~\cite{DBLP:journals/corr/abs-2203-11647}, considering both action and visual information.
As shown in Tab.~\ref{tab:action_diversity}, we measure complexity and diversity for action information.
For complexity, we calculate the variance of positions and rotations across different time intervals within each action sequence. Then, we take the mean across all videos to measure the dataset's \textit{average complexity}.
To measure the \textit{average diversity} of action sequences, we uniformly sampled each sequence for 300 time ticks across datasets. The variance across sequences at individual time points was then calculated and averaged over the sequence duration to determine the extent of variation among different action sequences.

For visual information (Tab.~\ref{tab:video_diversity}), we extract features from each video using a
pre-trained I3D model\cite{DBLP:journals/ijcv/GuptaTAMTDS21} after standardizing to 256 frames across sequences for all datasets. We then calculate the global standard deviation of these features to measure diversity. 

In Tables ~\ref{tab:action_diversity} and~\ref{tab:video_diversity}, a lower value means the dataset is less complex and diverse. For action diversity, both the human and robot demonstration datasets exhibit significantly higher position standard deviation than other datasets. Specifically, the position standard deviation for average complexity in human demonstrations (Tab. ~\ref{tab:action_diversity}) surpasses that of RoboTurk by $133.33\%$ and MIME by $3219.51\%$. Similarly, the rotation standard deviation (Euler angle) for average complexity in human-controlled robot demonstrations exceeds that of RoboTurk by $14.63\%$ and MIME by $38.56\%$. This trend can also be observed for visual diversity, where the standard deviation in our human demonstration dataset is $7.91\%$ greater than that observed in VideoFolding for comparable tasks. Similarly, our human-controlled robot dataset's standard deviation exceeds RoboTurk's by $5.33\%$. We can, therefore, conclude that Flat'n'Fold shows a broader range of visual and action data and demonstrates the applicability of our dataset in diverse settings for garment perception and manipulation.

\begin{table}[t]
\centering
\caption{Comparative analysis for video diversity}
\label{tab:video_diversity}
\begin{tabular}{@{}lcc@{}}
\toprule
\textbf{Dataset} & \textbf{Human} & \textbf{Robot} \\\midrule
Roboturk~\cite{DBLP:conf/iros/MandlekarBSTGZG19} & - & 589.99 \\
MIME~\cite{DBLP:conf/corl/SharmaMPG18} & 377.65 & 473.12  \\
Videofolding~\cite{DBLP:journals/ijrr/VerleysenBw20} & 688.07 & -\\
Semantic state est.~\cite{DBLP:journals/corr/abs-2203-11647} & 428.10 & 375.77\\
Ours & 742.53 & 621.45 \\
\bottomrule
\end{tabular}
\end{table}


\subsection{Grasping Point Prediction Benchmark}\label{sec:grasp_prediction}

Grasping point prediction is an essential subtask for garment manipulation but remains challenging in part due to a lack of comprehensive training datasets with annotated ground truth.
Research into garment grasping points has been confined mainly to simplistic scenarios, such as picking up centrally located points on crumpled fabrics, garments on a flat surface \cite{DBLP:conf/icar/ShehawyRZ21} or grasping garments hung on hangers \cite{DBLP:conf/icra/0030D20}. 
Our dataset addresses this gap by providing annotated grasping points while manipulating garments from a crumpled state to a folded state. 

To establish a benchmark for subsequent studies, we define a subset of our dataset comprising points in time when the hand/gripper is about to grasp the garment. We extract ground truth information at the instant of grasping, i.e. which hand grasps the garment (left/right), its position and its rotation. We extract point-clouds from 8 to 10 frames earlier, fusing the three views and segmenting to include only the garment; this avoids leakage of the arm position for grasping prediction. 
This yields 6,329 annotated point clouds from the human subset and 5,574 from the robot subset. The goal is then to predict the optimal grasp location given a point-cloud. As metrics, we measure the classification accuracy (left vs right hand), 
the L1 error of positions and the geodesic error of rotations. The dataset is divided into training, validation, and testing sets in a 7:1:2 ratio across all garment types. 
As baselines, we evaluate two popular models operating on point-clouds, PointNet++~\cite{DBLP:conf/nips/QiYSG17} and Point-BERT~\cite{DBLP:conf/cvpr/YuTR00L22}.
Both were pretrained on ModelNet-40, and we added two fully-connected layers to predict hand position (supervised with an L1 loss), rotation quaternion (with geodesic loss), and left/right hand (with cross-entropy loss).
We conducted two experiments using these models to examine the effects of varying dataset sizes and garment types on the outcomes.

In Tab.~\ref{tab:grasp_point_random}, we show results when training on different fractions of our dataset. All metrics consistently improve with the amount of training data in human and robot demonstrations, which shows the value of using a large-scale dataset for grasping garments. For example, on the human demonstrations subset, PointNet++'s classification accuracy increases from $55.7\%$ using $20\%$ of training data to $59.9\%$ with the full training set, with corresponding reductions in position error (from $0.103$ meters to $0.1$ meters) and rotation error (from $0.028$ radians to $0.023$ radians). 
However, even with the full dataset, the results are still far from perfect which indicates that existing methods struggle to predict grasps accurately and that future research has significant room for improvement. By comparing the two subsets, robot demonstrations outperform human demonstrations with higher classification accuracy and lower position errors across both methods but with higher rotation errors. The difference lies in the fact that robot demonstrations have a wider range of variation in the rotation component compared to human demonstrations (see Tab. \ref{tab:action_diversity}), causing the baseline models to generate higher rotation errors. Notably, for position errors, where robot demonstrations show less diversity, the errors are smaller.



%
In Tab.~\ref{tab:grasp_point_garment}, we show results split by garment type. In the table, \textit{Class. Acc.} is Classification Accuracy, \textit{Pos. Error} is Position Error, and \textit{Rot. Error} is the Rotation Error. \textit{H} represents Human demonstrations and `R' Robot demonstrations. For human demonstrations, we can observe that shirts (SS-shirt and SS-tshirt) have higher classification accuracy and lower rotation errors but higher position errors. In contrast, napkins have higher position errors but higher rotation errors. Pants, on the other hand, show lower classification accuracy. In robot demonstrations, towels have both higher position and rotation errors. These results indicate that these four garment types should be a particular focus for future work.

\begin{figure*}[t]
    \centering
    \includegraphics[width=\linewidth]{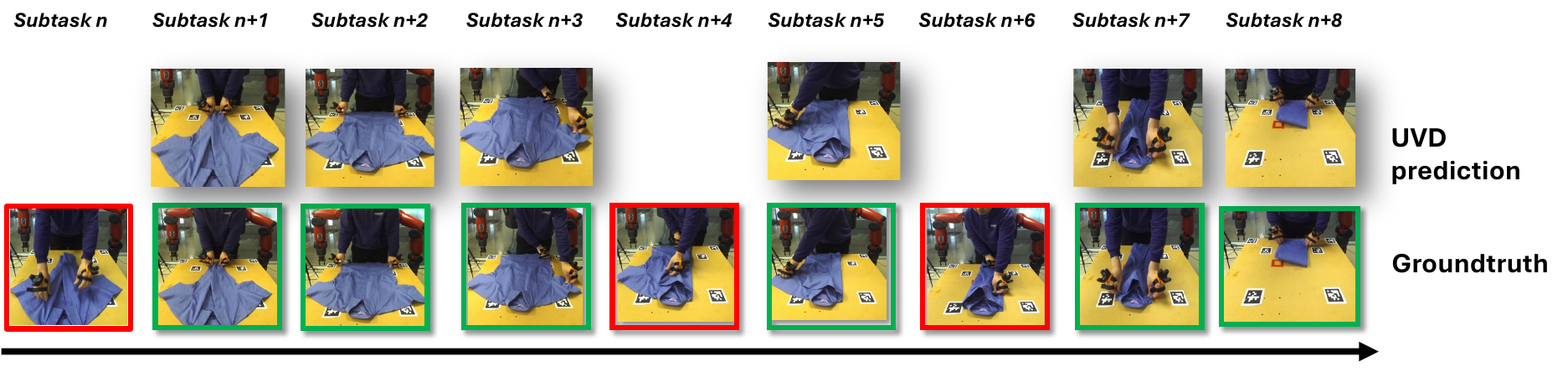}
    \vspace{-0.5cm}
    \caption{Example of subtask decomposition with Flat'n'Fold using UVD~\cite{zhang2024universal} alongside ground truth comparisons. Green blocks represent accurately predicted subtasks by UVD, while red blocks indicate ground truth subtask images that UVD failed to identify.}
    \label{fig:subtaskexample}
\end{figure*}

\begin{table}[t]
\centering
\caption{Grasping Point Prediction with PointNet++ and Point-BERT} 
\label{tab:grasp_point_random}
\resizebox{0.48\textwidth}{!}{
\begin{tabular}{@{}lccccccc@{}}
\toprule
\multirow{2}{*}{\textbf{Model}} & \multirow{2}{*}{\textbf{Dataset Size}}  &\multicolumn{2}{c}{\textbf{Classification acc.}} & \multicolumn{2}{c}{\textbf{Position error}}& \multicolumn{2}{c}{\textbf{Rotation error}} \\
\cmidrule{3-4}\cmidrule(l){5-6}\cmidrule(l){7-8}
& & Human & Robot & Human & Robot & Human & Robot \\\midrule
\multirow{3}{*}{PointNet++~\cite{DBLP:conf/nips/QiYSG17}} & 20\% & 0.557 & 0.669 & 0.103 & 0.092 & 0.028 & 0.048 \\
 & 50\% & 0.586 & 0.707 & 0.102 & 0.084 & 0.025 & 0.047 \\
 & 100\% & 0.599 & 0.710 & 0.100 & 0.076 &0.023  & 0.043 \\
\midrule
\multirow{3}{*}{Point-BERT~\cite{DBLP:conf/cvpr/YuTR00L22}} & 20\% & 0.606 & 0.658  & 0.101 & 0.073 & 0.023 & 0.049 \\
 & 50\% & 0.620 & 0.683 & 0.103 & 0.072 & 0.022 & 0.048 \\
 & 100\% & 0.625 & 0.747 & 0.096 & 0.067 &0.022  & 0.045 \\
\bottomrule
\end{tabular}
}

\vspace{2pt}
\hfill \textit{* Position error is given in meters and rotation error in radians}~~
\end{table}

\begin{table}[t]
\caption{Performance comparison of PointNet++ and Point-BERT}
\label{tab:grasp_point_garment}
\centering
\resizebox{0.48\textwidth}{!}{
\begin{tabular}{@{}l cccc cccc cccc@{}}
\toprule
 & \multicolumn{6}{c}{\textbf{PointNet++}} & \multicolumn{6}{c}{\textbf{Point-BERT}} \\ 
\cmidrule(r){2-7}\cmidrule(l){8-13}
 & \multicolumn{2}{c}{\textbf{Class. Acc.}} & \multicolumn{2}{c}{\textbf{Pos. Error}} & \multicolumn{2}{c}{\textbf{Rot. Error}} & \multicolumn{2}{c}{\textbf{Class. Acc.}} & \multicolumn{2}{c}{\textbf{Pos. Error}} & \multicolumn{2}{c}{\textbf{Rot. Error}} \\ 
\cmidrule(r){2-3}\cmidrule(r){4-5}\cmidrule(r){6-7} \cmidrule(lr){8-9}\cmidrule(r){10-11}\cmidrule{12-13}
 & H & R& H & R & H & R & H & R & H& R & H& R \\
\midrule
Napkin & 0.590 & 0.850 & 0.085 & 0.078 & 0.036 & 0.048 & 0.532 & 0.897& 0.065 & 0.071 & 0.039 & 0.040\\
Towel & 0.606 & 0.647 & 0.091 & 0.102 & 0.027 & 0.074 & 0.634 & 0.686 & 0.093 & 0.083 & 0.028 & 0.076 \\
SS-shirt & 0.661 & 0.747 & 0.118 & 0.070 & 0.019 & 0.050 & 0.719 & 0.635 & 0.093 & 0.061 & 0.016 & 0.060 \\
SS-tshirt & 0.657 & 0.671 & 0.121 & 0.067 & 0.023 & 0.053 & 0.729 & 0.671 & 0.108 & 0.053 & 0.021 &  0.061\\
LS-shirt & 0.549 & 0.667 & 0.102 & 0.083 & 0.023 & 0.056 & 0.576 & 0.648& 0.086 & 0.069 & 0.021 & 0.052 \\
LS-tshirt & 0.633 & 0.630 & 0.093 & 0.076 & 0.024 & 0.042 & 0.622 & 0.674 & 0.080 & 0.081 & 0.025 & 0.065 \\
Sweater & 0.606 & 0.746 & 0.091 & 0.078  & 0.027 & 0.028 & 0.592 & 0.651 & 0.076 & 0.071 & 0.023 & 0.042 \\
Pant & 0.558 & 0.776 & 0.101 & 0.070 & 0.035 & 0.052 & 0.475 & 0.834 & 0.087 & 0.065 & 0.038 & 0.039 \\ \midrule
Average &0.607 &0.717 &0.1 &0.078
&0.027 &0.05 & 0.61& 0.712&0.086 &0.069 &  0.026 & 0.054 \\
\bottomrule
\end{tabular}
}

\vspace{2pt}
\hfill\textit{* Position error is given in meters and rotation error in radians}~~
\end{table}

\subsection{Automated Subtask Decomposition Benchmark}
\label{sec:subtask_decomposition}

\begin{table}[] 
\centering 
\caption{Quantitative results for UVD compared across human and robot datasets and flattening/folding phases}
\label{tab:UVD_total}
\begin{tabular}{@{}lcccc} 
\toprule 
 & & \textbf{Precision} & \textbf{Recall} & \textbf{F1} \\
\midrule
\multirow{3}{*}{\textbf{Human}} & Flattening& 0.520 & 0.658 & 0.564\\
 & Folding & 0.905 & 0.582 & 0.695 \\
 & Total & 0.713 & 0.620 & 0.629 \\ \midrule
\multirow{3}{*}{\textbf{Robot}} & Flattening & 0.416 & 0.331 & 0.356 \\
 & Folding & 0.825 & 0.391 & 0.523 \\
 & Total & 0.621 & 0.361  & 0.441 \\
\bottomrule 
\end{tabular}
\end{table}

\begin{table}[t]
\centering
\caption{Quantitative results for UVD \cite{zhang2024universal} on human videos of different folding strategies}
\begin{tabular}{@{}lcccc@{}}
\toprule
\textbf{Strategy} & \textbf{Precision} & \textbf{Recall}  & \textbf{F1} \\\midrule
Fixed & 0.905 & 0.582 & 0.695 \\
Daily-life & 0.894 & 0.571  & 0.684 \\
\bottomrule
\end{tabular}
\label{tab:UVD}
\end{table}

Mastering long-horizon manipulation tasks such as flattening and folding remains challenging. One approach is to break them into smaller, more manageable subtasks to effectively learn these tasks and generalize to new situations.
We, therefore, use our dataset to define a benchmark for task decomposition methods; an example of task decomposition can be found in Fig. \ref{fig:subtaskexample}.



We define `\textit{pick}' and `\textit{place}' actions  ground-truth sub-task boundaries. Specifically, we divide each image sequence into two phases. During the \textit{flattening} phase, we treat every pick action as a subtask but not place since these often cause no change to the garment. During the \textit{folding} phase, we treat every pick or place as a subtask.
During the evaluation, we consider a predicted subtask boundary correct if it is within 10 frames of the annotated ground truth; based on this, we calculate the precision, recall, and F1 score.

As a baseline, we evaluate the state-of-the-art unsupervised task decomposition UVD~\cite{zhang2024universal}, using VIP~\cite{ma2022vip} as the visual encoder 
Results are shown in Tab.~\ref{tab:UVD_total}. The precision for both human demonstration and robot demonstration is relatively high at $0.713$ and $0.621$, respectively. This shows that the method accurately identifies relevant subtasks when the actions involve direct manipulation on a stable surface. However, lower recall rates of $0.620$ for humans and $0.361$ indicate that many subtask boundaries are missed. For both subsets, precision varied across phases, with a lower rate of $0.520$ and $0.416$ during flattening but a much higher rate of $0.905$ and $0.825$ during folding. When comparing the two subsets, the UVD method demonstrates better performance across all metrics in human demonstrations.

We also further analyzed whether different folding strategies affect the effectiveness of UVD. The results from Tab.~\ref{tab:UVD} show that varied folding approaches introduce complexities that affect the UVD method's performance. Specifically, participants folding garments in their style (Daily-life) demonstrated lower precision ($0.894$) and recall ($0.571$) compared to those following predefined rules (Fixed), which achieved a precision of $0.905$ and a recall of $0.852$, resulting in an F1 score of $0.684$. The latter indicates that there is a need to develop approaches that can capture more diverse manipulation strategies.



\section{Conclusion}
We have introduced Flat'n'Fold, a dataset comprising over 1,212 human and 887 robot demonstrations of flattening and folding 44 unique garments spanning eight categories. Flat'n'Fold contains high-resolution multi-view RGB-D and point cloud data, as well as ground-truth actions. We have also defined two new benchmarks for garment manipulation tasks: grasping point prediction and subtask decomposition.
he future, the size and scope of our dataset can enable training and benchmarking on new tasks. By providing diverse human actions, the dataset can be used for imitation learning, enabling robots to learn complex manipulation tasks. Additionally, Flat'n'Fold's multimodality can support the future development of models that accurately perceive and predict the state of garments, which is essential for improving the performance of cloth pose estimation and planning manipulation tasks.

\section*{ACKNOWLEDGMENT}
We want to thank Zhuo He, Tanatta Chaichakan, and the Computer Vision and Autonomous Systems (CVAS) research group for insightful discussions and for participating in the data collection for this work.



\bibliographystyle{IEEEtran}
\bibliography{References.bib}

\end{document}